\documentclass[10pt,twocolumn,letterpaper]{article}

\usepackage{cvpr}              
\usepackage{amsmath}
\usepackage{mdframed}
\usepackage{enumitem}
\usepackage{algorithm}
\usepackage{algorithmic}

\usepackage[T1]{fontenc}
\usepackage[section]{placeins}

\usepackage[scaled]{beramono}  
\usepackage{makecell}
\usepackage{booktabs}
\usepackage{array}
\usepackage{underscore}  
\usepackage{url}         
\usepackage{amsmath}
\usepackage{tikz}
\usepackage{stfloats} 
\usepackage{cuted} 
\usepackage{xcolor}
\usepackage[table,xcdraw]{xcolor}  
\usepackage{colortbl}               
\usepackage{multirow}               
\usepackage{graphicx}               

\definecolor{lavender}{HTML}{EEE9F3}
\definecolor{lavenderDeep}{HTML}{D6C9E6}

\usepackage{tcolorbox}
\usepackage{listings}

\newtcbox{\capbox}{on line,
  colframe=blue!50!black,
  colback=blue!3!white,
  boxrule=0.7pt,
  arc=2pt,
  boxsep=1pt,
  left=1pt, right=1pt, top=1pt, bottom=1pt,
  tcbox raise base,
  fontupper=\ttfamily\bfseries\small
}

\tcbset{
  capboxstyle/.style={
    on line,
    boxrule=0.7pt,
    arc=2pt,
    boxsep=1pt,
    left=1pt, right=1pt, top=1pt, bottom=1pt,
    tcbox raise base,
    fontupper=\ttfamily\bfseries\small
  }
}

\newtcbox{\capboxblue}{capboxstyle,
  colframe=blue!50!black,
  colback=blue!3!white
}

\newtcbox{\capboxgreen}{capboxstyle,
  colframe=green!50!black,
  colback=green!3!white
}

\newtcbox{\capboxred}{capboxstyle,
  colframe=red!50!black,
  colback=red!3!white
}

\newtcbox{\capboxpurple}{capboxstyle,
  colframe=purple!50!black,
  colback=purple!3!white
}

\newtcbox{\capboxorange}{capboxstyle,
  colframe=orange!60!black,
  colback=orange!5!white
}

\newtcbox{\capboxteal}{capboxstyle,
  colframe=teal!60!black,
  colback=teal!5!white
}

\definecolor{capiblue}{RGB}{232,241,250}
\definecolor{capigreen}{RGB}{232,248,240}
\definecolor{capipink}{RGB}{250,236,240}
\mdfsetup{skipabove=0pt,skipbelow=0pt}

\definecolor{cvprblue}{rgb}{0.21,0.49,0.74}
\usepackage[pagebackref,breaklinks,colorlinks,allcolors=cvprblue]{hyperref}

\def\paperID{22718} 
\def\confName{CVPR}
\def\confYear{2026}

\title{
\hspace{2em}%
\raisebox{-0.28\height}{%
  \includegraphics[height=2.1em]{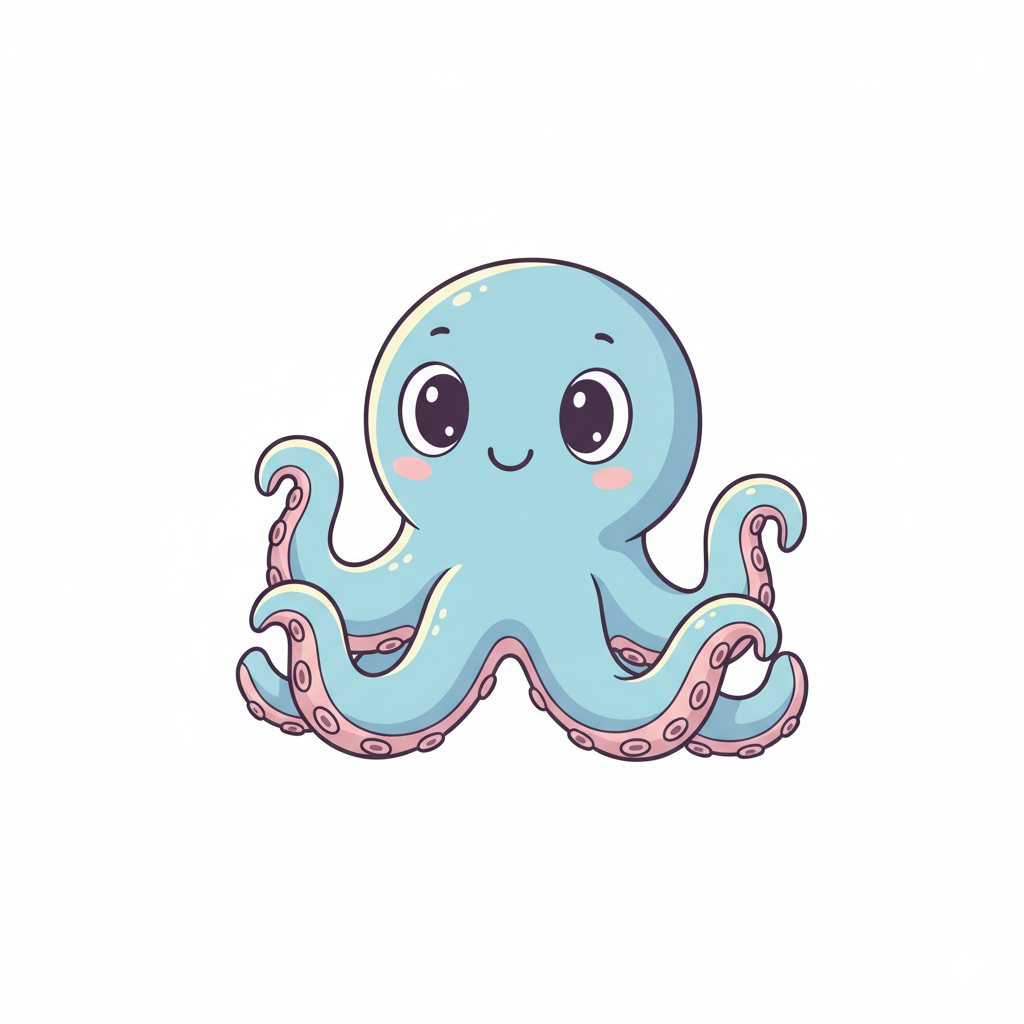}%
}\hspace{-0.5em}
Octopus: Agentic Multimodal Reasoning with Six-Capability Orchestration
}

\author{
\textbf{
Yifu Guo$^{1*}$ \quad
Zishan Xu$^{2*}$ \quad
Zhiyuan Yao$^{3*}$ \quad
Yuquan Lu$^{1}$ \quad
Jiaye Lin$^{4}$
}\\[4pt]
\textbf{
Sen Hu$^{5}$ \quad
Zhenheng Tang$^{6}$ \quad
Huacan Wang$^{7\dagger}$ \quad
Ronghao Chen$^{5\dagger}$
}
\\[10pt]
$^{1}$Sun Yat-sen University \quad
$^{2}$Shanghai Jiao Tong University \quad
$^{3}$Zhejiang University \\
$^{4}$Tsinghua University \quad
$^{5}$Peking University \quad
$^{6}$The Hong Kong University of Science and Technology \\
$^{7}$University of Chinese Academy of Sciences
\\[6pt]
\small $^{*}$Equal contribution \quad $^{\dagger}$Corresponding authors
}

\begin{document}

\maketitle


\begin{strip}
\begin{center}
\vspace{-2cm}
\includegraphics[width=\linewidth]{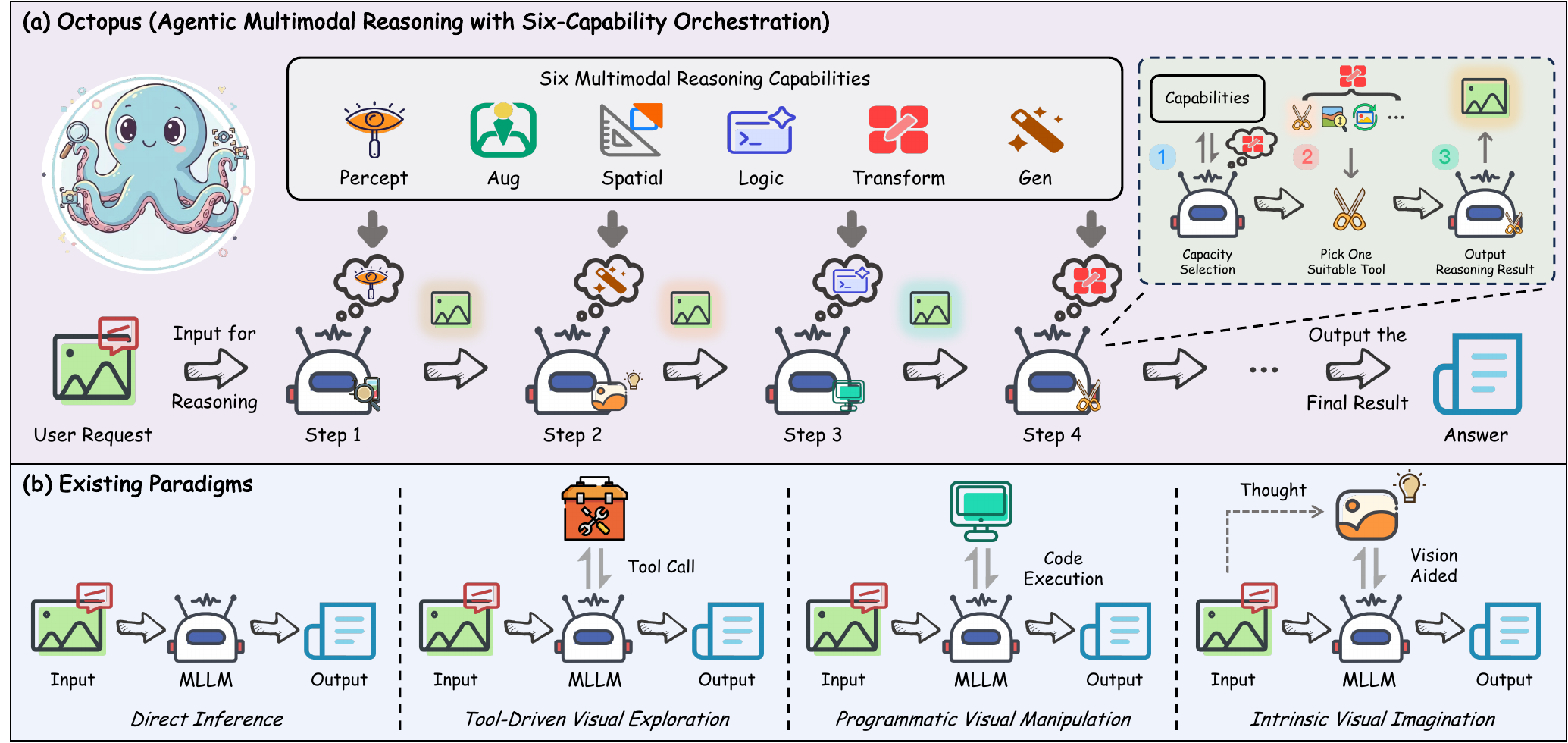}
\captionof{figure}{Overview of (a) our proposed \textbf{Octopus} and (b) existing multimodal reasoning paradigms.
Octopus dynamically selects one of the six core capabilities at each reasoning step and invokes its corresponding tool to solve the subproblem, enabling flexible and adaptive multimodal reasoning.}
\label{fig:framework_comparison}
\vspace{-0.3cm}
\end{center}
\end{strip}


\begin{abstract}

Existing multimodal reasoning models and frameworks suffer from fundamental architectural limitations: most lack the human-like ability to autonomously explore diverse reasoning pathways—whether in direct inference, tool-driven visual exploration, programmatic visual manipulation, or intrinsic visual imagination. Consequently, they struggle to adapt to dynamically changing capability requirements in real-world tasks. Meanwhile, humans exhibit a complementary set of thinking abilities when addressing such tasks, whereas existing methods typically cover only a subset of these dimensions.
Inspired by this, we propose Octopus: Agentic Multimodal Reasoning with Six-Capability Orchestration, a new paradigm for multimodal agentic reasoning. We define six core capabilities essential for multimodal reasoning and organize a comprehensive evaluation benchmark, Octopus-Bench, accordingly. Octopus is capable of autonomously exploring during reasoning and dynamically selecting the most appropriate capability based on the current state. Experimental results show that Octopus achieves the best performance on the vast majority of tasks in Octopus-Bench, highlighting the crucial role of capability coordination in agentic multimodal reasoning.

\end{abstract}
    
\section{Introduction}
\label{sec:intro}
In recent years, multimodal large language models (MLLMs) have exhibited rapid progress across a diverse range of tasks, largely driven by advances in step-by-step reasoning, mathematical derivation, and structured planning within language models~\cite{wei2022chain, yao2022react, kojima2022large, openai2024o1,du2025graphmaster,du2025mokgr,du2025graphoracle,wang2025repomasterautonomousexplorationunderstanding,luo2025codetestcasesenough}. As illustrated in~\Cref{fig:framework_comparison}(b), existing multimodal reasoning systems have evolved along several representative paradigms. Direct inference using MLLMs~\cite{liu2023visual, zhu2023minigpt, chen2023shikra} typically treats the visual information as static during reasoning, lacking mechanisms for active manipulation or iterative refinement. Tool-driven and programmatic approaches~\cite{suris2023vipergpt} partially address this limitation by employing predefined tools to solve decomposed subproblems. However, their performance is constrained by the limited scale of available tools, the lack of systematic organization, and the challenges associated with debugging and validating generated code. Meanwhile, internal visual imagination methods~\cite{chern2025thinking} aim to enhance the model’s exploratory capacity, but in the absence of effective coordination with other reasoning capabilities, their applicability remains restricted to specific domains such as path planning.

In stark contrast, humans demonstrate not only strong autonomy in visual reasoning tasks~\cite{vaishnav2025cognitive} but also a rich repertoire of cognitive capabilities that jointly support effective problem solving. Human reasoning is inherently multi-faceted: individuals flexibly engage perception, spatial understanding, logical deduction, abstraction, and imagination depending on the evolving needs of the task. This means that human problem solving is governed not by a single, universal mechanism, but by the dynamic orchestration of diverse capabilities.

Such autonomy manifests in the ability to adaptively select, combine, and refine capabilities across different stages of a task. For example, when tackling a visual problem, a person may begin by identifying fine-grained visual elements (e.g., endpoints or obstacles), then infer geometric or path constraints, and subsequently use visual annotation or mental simulation to validate ambiguous steps. The essence of human visual reasoning thus lies in the interplay between self-directed decision-making and the strategic deployment of distinct cognitive abilities~\cite{kunda2013computational}. Large reasoning models (LRMs), trained through autonomous reinforcement-style procedures, have recently shown emergent behaviors such as self-verification, reflection, and long-chain reasoning~\cite{seagent}. These behaviors mirror, at a coarse level, the agentic and capability-driven processes seen in humans—precisely the type of unified mechanism we aim to instill in multimodal agents.


Building on recent agentic reasoning frameworks such as Search-o1~\cite{li2025searcho1agenticsearchenhancedlarge}, we introduce \textbf{Octopus}: Agentic Multimodal Reasoning with Six-Capability Orchestration, an agentic framework designed to operationalize this human-like flexibility, as illustrated in ~\Cref{fig:framework_comparison}(a). Octopus allows a model to dynamically invoke and compose different visual and cognitive abilities along the reasoning trajectory. To achieve this, we first systematically decompose multimodal reasoning into six fundamental capabilities. 
For example, Octopus can leverage its perceptual capability to capture visual information, employ the augmentation capability to highlight and annotate images, and utilize the generative capability to maintain its internal imagination space. Based on this decomposition, we build our framework and organize a comprehensive benchmark to evaluate these crucial agentic multimodal reasoning abilities.


In summary, our main contributions are as follows:





\begin{itemize}
    \item  We systematically define six fundamental capabilities that underlie multimodal reasoning, and on this basis, we construct a comprehensive benchmark to evaluate the capability of multimodal large models and agentic systems.

    \item We propose a capability-based multimodal agentic reasoning paradigm that emphasizes autonomous switching between different capabilities along the reasoning trajectory.
    \item On our curated benchmark, our approach achieves the best performance on most tasks, highlighting the critical role of capability Orchestration in enhancing agentic reasoning.
\end{itemize}

\section{Related Work}
\label{sec:Related}

\begin{figure*}[t!]
    \centering
    \includegraphics[width=\textwidth]{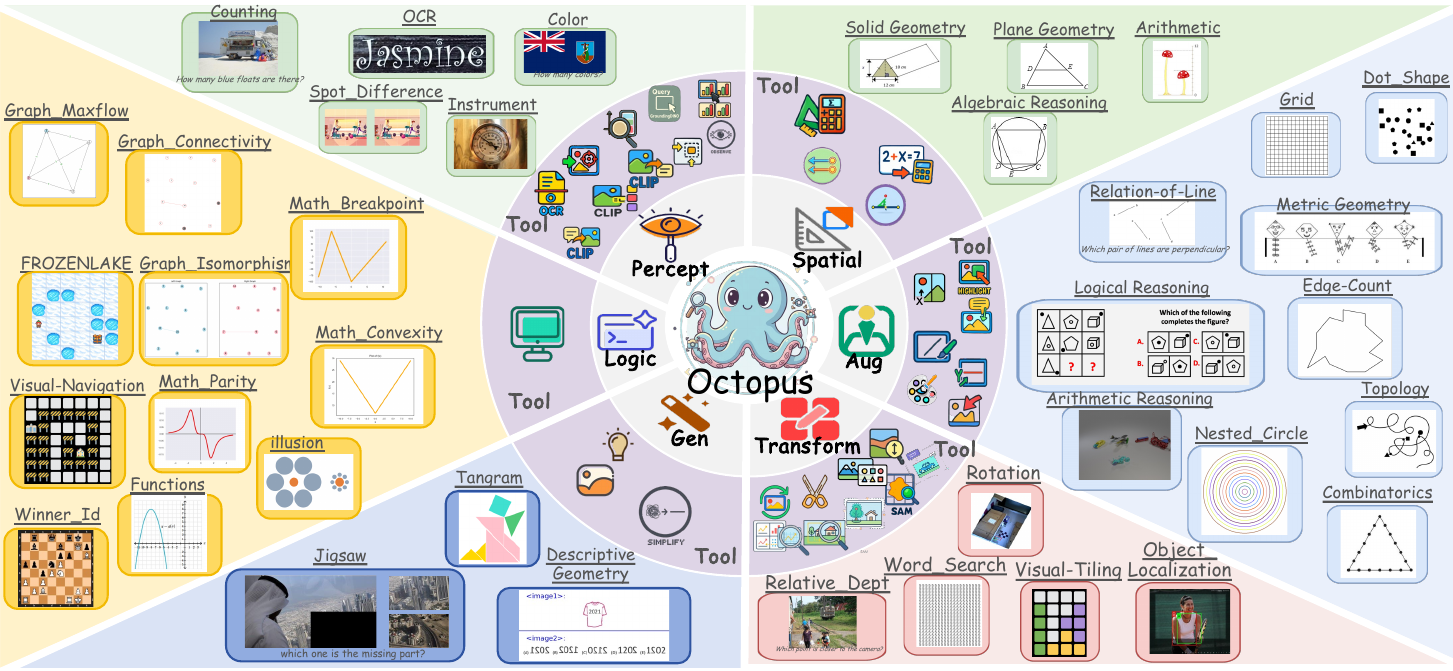}
    \caption{\textbf{Capability-centric organization of Octopus-Bench}. This benchmark enables comprehensive assessment of multimodal reasoning capability. The innermost circle represents the six fundamental multimodal reasoning capabilities. The middle ring contains the corresponding tool modules that operationalize each capability. 
The outer ring shows the diverse sub-tasks in our benchmark used to evaluate each capability dimension.}
    \label{fig:placeholder}
\end{figure*}
\paragraph{Direct Inference by Multimodal/Vision LLMs.}
Recent large vision–language models (LVLMs) integrate early-stage visual encoders and language models through joint training to form more comprehensive multimodal reasoning systems. 
Models such as GPT-4V~\cite{yang2023dawnlmmspreliminaryexplorations}, Gemini 2.5~\cite{comanici2025gemini}, Qwen2.5-VL~\cite{bai2025qwen2},LLaVA-OneVision~\cite{li2024llavaonevisioneasyvisualtask} and the InternVL~\cite{chen2024internvl,gao2024mini} series have demonstrated strong unified cross-modal understanding capabilities. 
More recently, several studies have extended the Chain-of-Thought (CoT)~\cite{wei2022chain} paradigm to multimodal contexts, prompting models to enhance reasoning with both textual and visual inputs~\cite{zhang2023multimodal}. 
However, direct-reasoning models typically treat the image as a static, one-shot input, leaving the visual information neither actively manipulable nor dynamically updatable during inference. The reasoning process still occurs primarily within the linguistic domain, resulting in a substantial semantic gap between visual structures, spatial constraints, and symbolic reasoning. This limits the model’s performance on tasks requiring fine-grained visual operations or long-horizon visual reasoning.

\paragraph{Tool-Driven Visual Exploration.}
To improve visual reasoning, prior work often relies on predefined external tools—such as detectors, segmenters, OCR systems, zooming, cropping, or search strategies~\cite{hu2024visual,wu2024dettoolchain,zhou2024image,decoupleq}. The model typically acts as a controller that invokes these tools via prompts or plugins and builds reasoning cues from their outputs.
However, these approaches usually depend on small, functionally limited toolsets that lack systematic organization, and most follow a single-step invocation paradigm. Consequently, the model cannot iteratively integrate diverse tools during inference to construct richer reasoning trajectories.
In contrast, we develop a substantially larger and systematically organized tool ecosystem covering dozens of visual capabilities mapped to six core cognitive competencies. Our framework further supports multi-round, iterative tool composition, enabling agentic, multi-step visual reasoning beyond one-shot tool scheduling.

\paragraph{Programmatic Visual Manipulation.}
Some works~\cite{suris2023vipergpt,Rodriguez_2025_CVPR,awal2025webmmu} enable models to perform visual operations such as cropping, annotation, and geometric computation by generating executable code, offering greater flexibility than fixed tool calls. 
However, these methods typically rely on a single LMM to directly generate full programs, causing the overall reasoning quality to be heavily constrained by the model’s code-generation capability. When the generated code contains errors, the system lacks robust fallback mechanisms, often leading to complete reasoning failure. Moreover, these approaches generally rely solely on program generation for visual manipulation and struggle to leverage existing high-precision visual tools in a complementary manner.
In contrast, we treat code generation as one callable capability, implemented through dedicated code-generation modules, and decompose complex visual tasks into finer-grained substeps to improve execution reliability.

\paragraph{Intrinsic Visual Imagination.}
A recent line of work, including Thinking with Generated Images~\cite{chern2025thinking}, Visual Planning~\cite{xu2025visual} and Visualization-of-Thought~\cite{wu2024mind}, explores the use of image generation to emulate human spatial imagination. 
While promising, these approaches are often restricted to specific domains such as navigation path visualization or geometric structure synthesis. 
Moreover, their generated-image reasoning processes cannot be seamlessly integrated with other cognitive abilities, limiting their generalization across diverse multimodal reasoning scenarios.
In contrast, our framework treats visual creation as one of six core capabilities and orchestrates it jointly with spatial understanding, logical programming, and other abilities. This design ensures that visual generation functions not as an isolated stage but as an integrated component of the overall reasoning trajectory.
\section{Method}
\subsection{Capability Mechanisms}








In existing multimodal reasoning research, models typically rely on a limited set of tool sets or code-execution modules to enhance their visual understanding. However, such approaches remain limited in capability coverage and fail to comprehensively capture the essential processes underlying multimodal reasoning. To address this gap, we systematically define six fundamental capabilities that together form a complete visual reasoning capability space.

These six capability mechanisms decompose multimodal reasoning into a set of composable atomic skills. Multimodal tasks inherently require the integration of information from both modalities and dynamic switching between capabilities during reasoning. For instance, in a geometric proof task, humans rely on Spatial \& Geometric Understanding understand spatial relations and record symbolic derivations on draft. Similarly, our Octopus framework employs corresponding capability-specific tools to handle such problems in a human-like manner—using bounding-box annotation and distance computation to perceive geometric relationships, and applying logical programming for precise symbolic calculation.


\begin{itemize}
    \item \textbf{Fine-grained Visual Perception}: 
    Accurately extracts structured visual information—such as text, object locations, or local attributes—to ground downstream reasoning. 
    This includes capabilities like retrieving pixel-level cues or matching visual–text semantics (e.g., \texttt{OCR} for text extraction, \texttt{grounding\_dino} for object localization).

    \item \textbf{Visual Augmentation \& Marking}: 
    Externalizes intermediate reasoning by adding interpretable visual cues onto the image, helping the agent highlight salient evidence or clarify logical steps 
    (e.g., \texttt{highlight} for region emphasis, \texttt{arrow} for directional annotation).

    \item \textbf{Spatial \& Geometric Understanding}: 
    Performs reasoning over geometric structures, spatial relations, and topological constraints, which is essential for geometry problems or diagram interpretation 
    (e.g.,\texttt{geometry\_calculator} for computing areas and perimeters, \texttt{geom\_perp\_intersect} for computing perpendicular intersections).

    \item \textbf{Logical Programming Reasoning}: 
    Executes structured symbolic operations through programmatic logic, enabling precise computation or algorithmic reasoning beyond natural language 
    (e.g., using \texttt{code\_agent} to run mathematical expressions or implement geometric solvers).

    \item \textbf{Visual Transformation \& Editing}: 
    Modifies visual content to simplify problem structure or isolate relevant components, supporting stepwise multimodal reasoning 
    (e.g., \texttt{crop} to focus on regions of interest, \texttt{sam} for segmentation-based decomposition).

    \item \textbf{Visual Creation \& Generation}: 
    Produces new visual artifacts—such as sketches or simplified diagrams—that serve as intermediate reasoning aids or creative outputs 
    (e.g., \texttt{generate\_image} for synthetic image creation, \texttt{simplify\_image} for structural abstraction).
\end{itemize}


\subsection{Problem Formulation}



We study a challenging multimodal reasoning task, where solving a question
requires combining several kinds of reasoning over both the image and text
modalities. Formally, each instance provides two main inputs: a task instruction $Q_T$ and an image $I_{\text{input}}$.
The model operates within a unified multimodal reasoning space and generates a sequence of reasoning steps $R_i$ throughout the inference process. Under this framework, Octopus is endowed with six cognitive capabilities, drawn from the capability set that governs how the model selects and applies different forms of multimodal reasoning.
\begin{equation}
C_i \in \{C_{\text{percept}},\,C_{\text{aug}},\,C_{\text{spatial}},\,C_{\text{logic}},\,C_{\text{transform}},\,C_{\text{gen}}\},
\end{equation}
which collectively correspond to the six core dimensions of multimodal reasoning.

At the $i$-th step, the agent maintains a state $\mathcal{E}_i$,
which demonstrates the current visual and textual evidence.
The reasoning and capability histories up to step $i-1$ are
\begin{equation}
    R_{<i} = \{R_0, R_1, \dots, R_{i-1}\}, \qquad
    C_{<i} = \{C_0, C_1, \dots, C_{i-1}\},
\end{equation}
where $C_{<i}$ encodes the different solution strategies that the agent has
attempted so far.

Given the current state and history, the agent chooses the next reasoning
operator $R_i$ and its concrete action $a_i$ according to a policy $\pi$:
\begin{equation}
    R_i
    = \arg\max_{R} \;
      \pi\!\left(R \mid I_{\text{input}}, Q_T,
      C_{<i}, R_{<i}, \mathcal{E}_{i-1}\right).
\end{equation}
After extracting the selected capability $C_i$ from the reasoning trace $R_i$, the agent executes $C_i$ to update the multimodal state
\begin{equation}
    (C_i, \mathcal{E}_{i-1}) \longrightarrow \mathcal{E}_i .
\end{equation}
By iteratively applying this procedure within the multimodal reasoning space, the model integrates visual and textual evidence and progressively moves toward the final answer.

\subsection{Agentic Reasoning and Capability Orchestration}

Octopus demonstrates a significant advantage in its autonomous exploration of both inter-task  diversity. Unlike many existing agent methods ---which may perform well on specific task types but inherently constrain the exploration space---Octopus enables the model to flexibly adapt its reasoning strategies according to varying task requirements. From the perspective of the overall reasoning space, this autonomy allows Octopus to pursue diverse reasoning trajectories through dynamic combinations of capabilities.
\begin{algorithm}[t]
\caption{Agentic Reasoning with Capability Orchestration in Octopus}
\label{alg:inference_pipe}
\begin{algorithmic}[1]
\REQUIRE Task instruction $Q_T$, input image $I_{\text{input}}$, Multi-modal reasoning model $\mathcal{M}$


\STATE \textbf{State Initialization:}
\STATE $C \leftarrow \emptyset$, $R \leftarrow \emptyset$
\STATE $\mathcal{E} \leftarrow \{ I_{\text{input}},\, Q_T \}$, $i \leftarrow 0$


\STATE \textbf{Main Reasoning Loop}
\WHILE{$i < \text{max\_turn}$}
    \STATE $i \leftarrow i + 1$
    
    \STATE \textit{// Agentic Reasoning}
    \STATE $R_i \leftarrow \mathcal{M}(\mathcal{E}_{i-1}, C_{<i}, R_{<i}, Q_T)$
    \STATE $R \leftarrow R \cup \{R_i\}$ 
    
    \IF{$R_i$ ends with \capboxteal{</answer>}}
        \STATE \textbf{break}
    \ENDIF
    
    \STATE \textit{// Choose Capability}
    \STATE $C_i \leftarrow \text{Extract}(R_i,\capbox{<cap>},\capbox{</cap>})$    
    \STATE $C \leftarrow C \cup \{C_i\}$
    \STATE $Tool \leftarrow \text{Extract}(R_i,\capboxred{<tool\_call>},\capboxred{</tool\_call>})$
    \vspace{-10pt}
    \STATE \textit{// Update State}
    \STATE $obs \leftarrow \text{Excute}(tool)$
    
    
    \STATE $\mathcal{E} \leftarrow \mathcal{E}_{i-1} \cup \{obs\}$
    
\ENDWHILE

\RETURN $\text{Extract}(R,\capboxteal{<answer>},\capboxteal{</answer>})$

\end{algorithmic}
\end{algorithm}
As illustrated in \Cref{alg:inference_pipe}, Octopus performs agentic reasoning through dynamic capability orchestration. Concretely, during the main reasoning process, given the current state and the historical reasoning trajectory—including previously selected capabilities and intermediate results—Octopus first produces an internal thought segment, encapsulated within the special tokens \capboxgreen{<think>}...\capboxgreen{</think>}, which expresses its latent deliberation for the current step. Based on this internal reasoning, the model then analyzes the ongoing context to determine which capability is most appropriate for the current subproblem.The selected capability is explicitly marked within the reasoning chain $R_i$ using \capbox{<cap>}...\capbox{</cap>}, which first determines the cognitive skill the model intends to apply.
Based on this chosen capability, Octopus then selects an appropriate tool from the corresponding capability-aligned toolset. The specific tool is identified through the \capboxred{<tool_call>}...\capboxred{</tool_call>} annotation, and its execution output is incorporated back into the multimodal reasoning context for subsequent steps.

This two-stage procedure—first selecting a capability, then choosing a tool conditioned on that capability—parallels how humans decompose multimodal tasks: people typically determine the type of cognitive capability a problem requires, before selecting the concrete action that can best accomplish it.

Finally, when the model concludes the task, the final solution is emitted within the designated answer tokens \capboxgreen{<answer>}...\capboxgreen{</answer>}.


This process can be formally expressed as:
\begin{equation}
    R_i = \mathcal{M}\big(\mathcal{E}_{i-1}, C_{<i}, R_{<i}, Q_T\big),
\end{equation}
where $\mathcal{E}_{i-1}$ denotes the reasoning context up to step $i-1$, $C_{<i}$ and $R_{<i}$ represent the previously invoked capabilities and reasoning results, and $Q_T$ denotes the task objective.

\section{Experiment}


\subsection{Octopus-Bench}
\begin{table*}[ht]
\centering
\resizebox{\linewidth}{!}{%
\begin{tabular}{l|ccccccccccccccc}
\toprule
\textbf{Model} &
\textbf{Sim} &
\textbf{Fore} &
\textbf{IQ} &
\textbf{Refl} &
\textbf{Count} &
\textbf{Depth} &
\textbf{Spatial} &
\textbf{Jigsaw} &
\textbf{VisCorr} &
\textbf{SemCorr} &
\textbf{ArtStyle} &
\textbf{FunCorr} &
\textbf{Local} &
\textbf{MultiV} &
\textbf{Avg} \\
\midrule

\rowcolor{gray!10} \multicolumn{16}{c}{Closed-source MLLMs}\\
\midrule
GPT-4o~\cite{achiam2023gpt} &65.4  &79.55  &30  &38.8  &51.7  &74.2  &69.2  &55.3  &75.0  &54.0  &82.9  &39.2  &56.0  &60.2  &59.39  \\
Gemini-2.5-Pro~\cite{comanici2025gemini} &55.9  &45.5  &27.3  &46.3  &65.00  &50  &67.1  &54.0  &37.2  &22.1  &49.5  &32.3  &46.4  &41.4  &45.71  \\
Claude-3.5-Sonnet~\cite{anthropic2024claude35sonnet}&64.2 & 74.3 &  26.2&41.3  & 47.4 &53.2  &71.7  & 41.8 &59.7  &39.6  &76.8  & 31.5 &51.7  &49.2  &52.04  \\
\midrule

\rowcolor{gray!10} \multicolumn{16}{c}{Open-source MLLMs} \\
\midrule
Qwen2VL-7B~\cite{wang2024qwen2} &53.33  &38.64  &16.9  &33.58  &\underline{73.33}  &57.26  &79.72  &54.0  &33.72  &31.65  &51.28  &18.46  &54.1  &45.11  &45.79  \\
Qwen2.5-VL-72B~\cite{bai2025qwen2} & 59.2 & 71.3 &22.2  &\underline{49.2}  & 57.7 &62.8  & 79.6 &12.6  &25.4  & 32.6 & 41.4 & 29.7 & 49.6 & 56.4 &46.41  \\
Qwen2.5-VL-32B~\cite{bai2025qwen2} & 41.2 &62.4  &17.5  &41.6  &49.7  &60.5  & 65.7 & 8.6 & 20.5 & 27.5 & 35.3 & 21.5 & 39.5 & 41.2 &38.05  \\
Qwen2.5-VL-7B~\cite{bai2025qwen2} & 79.26 &34.85  & 8.1 &25.37  &65.00  &65.32  &83.22  &56.67  &40.12  &24.46  &58.97  &19.23  &41.8  &43.61  &46.14  \\
LLaVA-v1.5-7B~\cite{liu2023llavapluslearningusetools}&46.3  &28.0  &24.0  &36.6  &43.4  &52.4  &61.5  &11.3  &25.6  &23.0  &47.9  &21.5  &48.8  &49.6  &37.14  \\
LLaVA-v1.5-13B~\cite{liu2023llavapluslearningusetools} &46.3  &27.3  &28.0  &45.5  &50.0  &53.2  &67.8  &58.0  &29.1  &32.4  &47.9  &20.8  &47.2  &41.4  &42.49  \\
LLaVA-1.6-M-7B ~\cite{li2024llavanext-strong}& 21.2 &38.2  & 3.3 &29.4  &31.6  & 47.6 & 42.3&  20.4& 9.4 & 2.4 & 15.6 & 10.2 & 21.4 & 24.6 &22.69  \\
LLaVA-Next-72B~\cite{li2024llavanext-strong} & 47.1 &69.3  &15.4  &44.5  &52.2 &52.4  &71.5  &10.5 &21.9  &32.3  &32.2  &23.3  &42.5  &45.3  &40.03  \\

\midrule

\rowcolor{gray!10} \multicolumn{16}{c}{General Training Models} \\
\midrule
DeepEyes-7B~\cite{zheng2025deepeyesincentivizingthinkingimages} & 63.64 & 39.4 & 13.6 & 31.1 & 61.2 & 83.9 & \underline{87.2} & 56.8 & 60.5 & 38.2 & 41.4 &  21.9&  51.2& 51.6 &50.1  \\
DeepSketcher-7B~\cite{zhang2025deepsketcherinternalizingvisualmanipulation} & 58.24 &27.3  & 29.8 &  39.4&62.3  &61.3  & 86.3 & 54.1& 51.2 & 32.4 &  37.9&31.2  & \textbf{73.3} & \underline{60.7} &50.4  \\
VTS-V~\cite{VTS-V}& \underline{85.19} &71.21  & \underline{30.3}& 40.30 &67.50  & 79.84 & 85.31 & \underline{75.33} & 82.56 & 56.83 & 80.34 &53.08  &68.85  & 52.63 &66.4  \\

\rowcolor{gray!10} \multicolumn{16}{c}{Multimodal Reasoning Frameworks} \\
\midrule
GPT-4o + Sketchpad~\cite{hu2024visual} &84.2  &79  &22.8  &33.1  &66.7  &\underline{83.9}  &81.1  &70.7  &80.8  &\textbf{58.3}  &77.19  &42.1  &65.4  &45.6  &63.64  \\
GPT-4o + CoT~\cite{achiam2023gpt} &63.70  &62.88  &23.11  &41.04  &65.00  &73.39  &82.52  &62.0  &82.56  &57.55  &82.05  &\textbf{57.69}  &60.66  &53.38  &61.97  \\
GPT-4o + SoM~\cite{yang2023setofmarkpromptingunleashesextraordinary}&63.70  &  60.32&  20.65& 36.32 &43.33  &68.55  &76.22  &49.33  &83.72  &52.52  & 80.92 &47.69  &59.84  &56.40  &61.6  \\
GPT-4o + PyVision~\cite{zhao2025pyvisionagenticvisiondynamic} &  76.2&72.1  & 20.3 &34.2  &67.9  & 79.3 &82.1  &66.3  &79.1 &57.2  & 75.2& 40.2&63.2&46.2& 61.4\\

GPT-4o + MMFactory~\cite{fan2024mmfactoryuniversalsolutionsearch} &75.30  &\underline{84.80}  & 28.7 &35.10  &61.70  &79.84  &81.80  &75.30  &\textbf{85.50}  &\textbf{58.3}  &\underline{83.0}  &\underline{55.40}  &59.00  &60.20  &\underline{68.86}  \\

\rowcolor{lavenderDeep}GPT-4o + Octopus& \textbf{90.21} & \textbf{86.1} &  \textbf{34.1}& \textbf{51.3} &\textbf{75.3}  & \textbf{85.1} & \textbf{90.2} & \textbf{78.1} & \underline{84.42} &\underline{58.1}  & \textbf{85.2} & 54.2 &\underline{70.8}  &  \textbf{62.1}& \textbf{71.8} \\ 
\bottomrule

\end{tabular}
}
\caption{
Results on the \textbf{Octopus-BLINK}, which evaluates MLLMs on 14 fine-grained visual perception and reasoning categories.
We report accuracy (\%) for closed-source MLLMs, open-source MLLMs, multimodal reasoning frameworks, and generally trained models, along with the overall average (Avg).
}
\label{tab:BLINK_performance_comparison}
\end{table*}
We evaluate our method on Octopus-Bench, a capability-centric evaluation suite constructed by resampling and reorganizing instances from several widely used multimodal reasoning benchmarks. The overview of Octopus-Bench is depicted in Figure~\ref{fig:placeholder}. Its sources include BLINK~\cite{blink}, TIR-Bench~\cite{tirbench}, IsoBench~\cite{fu2024isobench}, Geometry3K~\cite{lu2021intergps}, MathVerse~\cite{zhang2024mathverse}, WeMath~\cite{qiao2024wemath}, Math-Vision~\cite{wang2024mathvision} and MathVista~\cite{lu2023mathvista}. We additionally conduct supplementary experiments on COMT~\cite{comt}, $V^*$Bench~\cite{vstar}, and MMVP~\cite{mmvp}.

To construct Octopus-Bench, we sample and reorganize instances from the above benchmarks and annotate each instance with one or more of our six capability dimensions: fine-grained perception, visual augmentation and marking, spatial and geometric understanding, logical programming reasoning, visual transformation and editing, and visual creation and generation. We further include several visual navigation and visual tiling tasks—such as FrozenLake-style visual navigation~\cite{MMI_FrozenLake}—to compensate for the under-representation of interactive, long-horizon scenarios in existing benchmarks. Octopus-Bench thus provides a unified capability-oriented lens for evaluating multimodal reasoning models, enabling consistent model comparison across capability dimensions as well as fine-grained analysis of how an agent allocates and orchestrates its six capabilities across tasks of varying difficulty.

\begin{figure}[t!]
    \centering
    \includegraphics[width=0.48\textwidth]{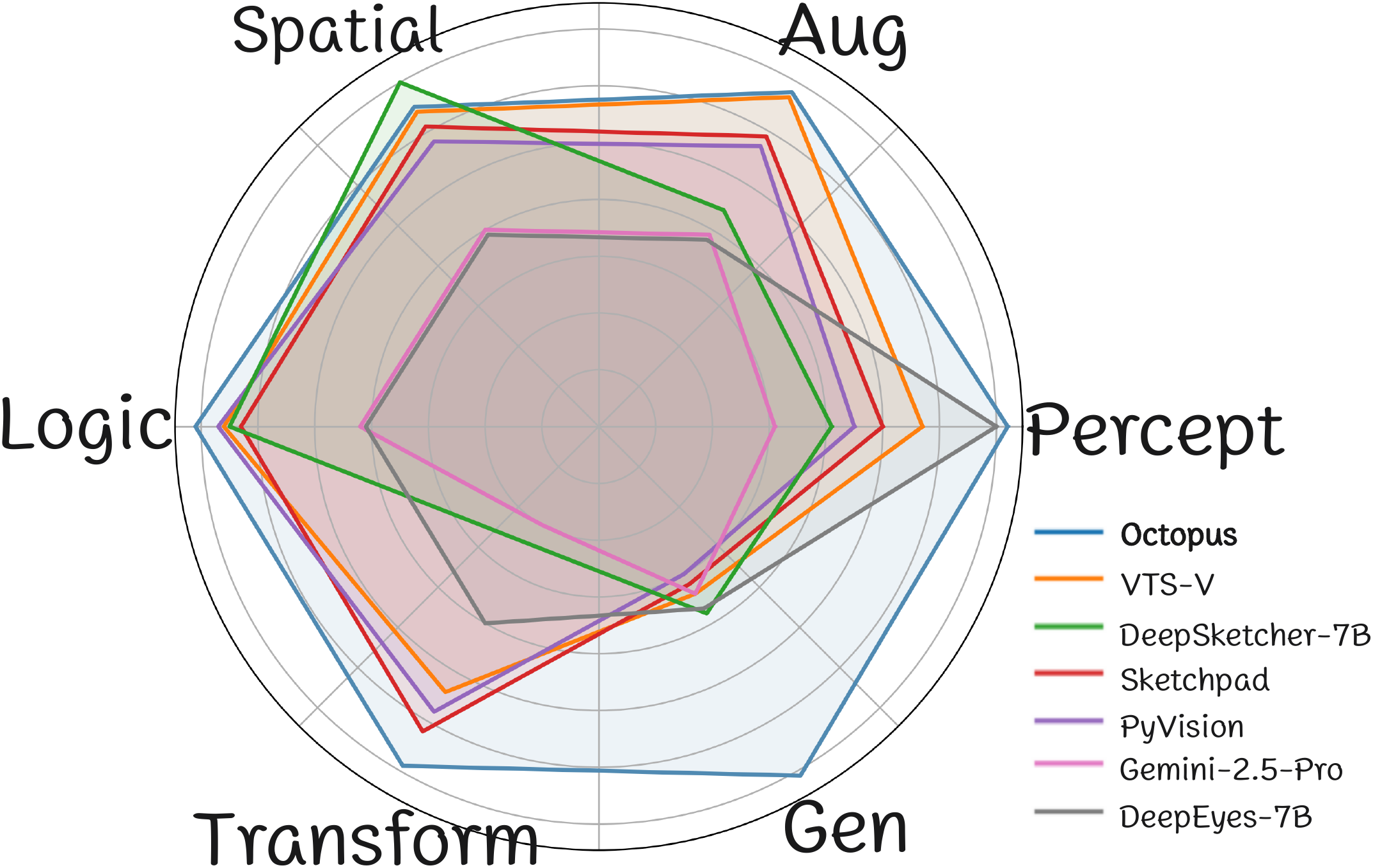}
    \caption{
\textbf{Capability comparison on Octopus-Bench}. Octopus shows the most balanced and consistently strong performance.
}
    \label{fig:capability_comparison}
\end{figure}


\subsection{Implementation Details}

\noindent\textbf{Backbone and Tool Models.}
We build Octopus on top of the off-the-shelf GPT-4o model~\cite{openai2024gpt4o}, which serves as the unified backbone for planning, capability selection, and high-level reasoning.
We use Claude 4.5 Sonnet~\cite{anthropic2025claudesonnet} as the primary driver for our code tool, serving as the core model for programmatic and symbolic reasoning.
For fine-grained visual perception, our observation tool is powered by Gemini 2.5 Flash~\cite{team2025gemini}, which we use for operations such as OCR, and region-level captioning.
All these models are accessed via their official APIs without any additional fine-tuning or task-specific training; our contributions lie entirely in the paradigm design and capability orchestration.
\begin{figure}
    \centering
    \includegraphics[width=1\linewidth]{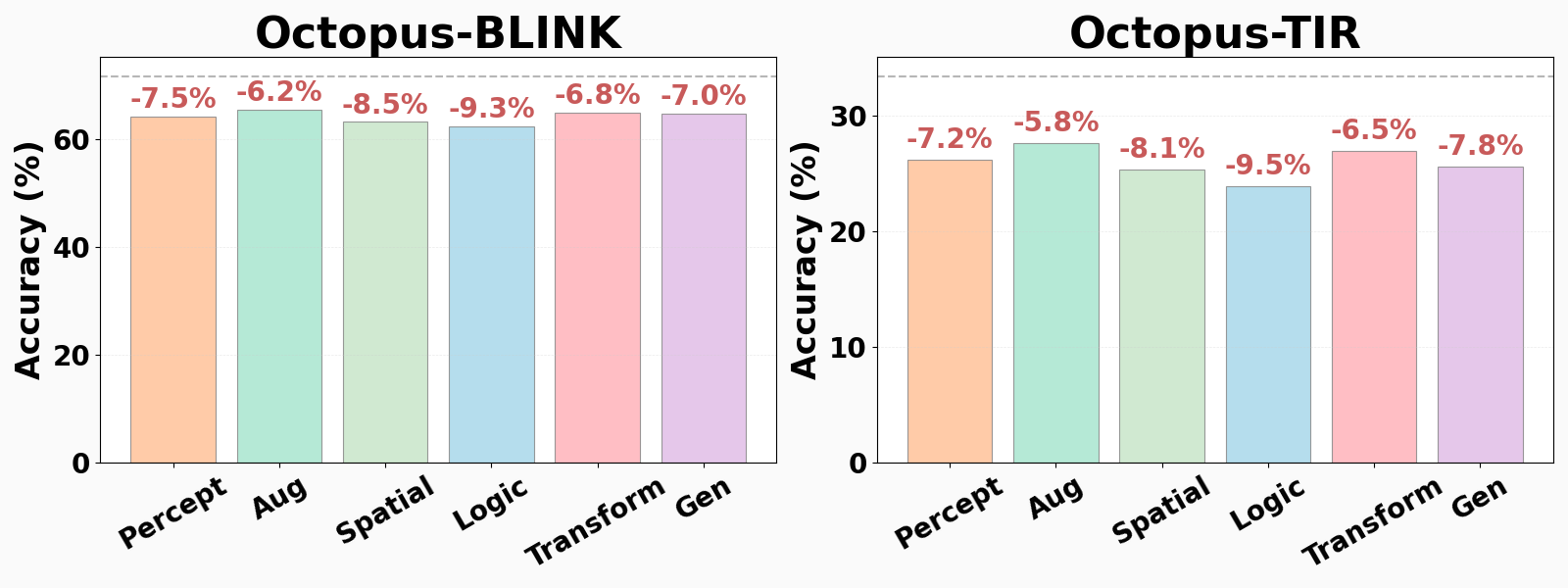}
    \caption{Ablation study of individual capabilities on Octopus-Blink and Octopus-TIR.}
    \label{fig:ablation_cap}
\end{figure}

\noindent\textbf{Capability and Tool Specification.}
We encode the six capability dimensions and their associated tools directly in the system prompt.
Each capability is given a concise natural-language description and a list of admissible tools, and the agent is instructed to (i) first select a capability, (ii) plan a sequence of tool calls within that capability, and (iii) summarize intermediate observations before producing the final answer.
This protocol is implemented via a unified prompting template with explicit markers for internal reasoning, capability selection, tool invocations, and final responses. 

\noindent\textbf{Inference Configuration.}
We use a unified inference configuration across all benchmarks and models.
We allow long-context reasoning by capping the effective context length at roughly $60\%$ of the maximum context window supported by GPT\mbox{-}4o, and we permit up to $10$ reasoning turns.
Decoding uses a temperature of $\tau=0.3$ and top-$p$ of $p=1.0$, yielding effectively near-deterministic behaviour across runs.
Closed-source baselines are run with their official or recommended settings whenever available, and we align decoding hyperparameters with our agent as much as possible to ensure a fair comparison.
\begin{table*}[ht]
\centering
\resizebox{\linewidth}{!}{%
\begin{tabular}{lcccccccccccccc}
\toprule
\textbf{Model} &
\textbf{All} &
\textbf{Color} &
\textbf{Pro} &
\textbf{OCR} &
\textbf{SR} &
\textbf{Maze} &
\textbf{Math} &
\textbf{WS} &
\textbf{LL-VQA} &
\textbf{IR} &
\textbf{SD} &
\textbf{JG} &
\textbf{VS} &
\textbf{RG} \\
\midrule

\rowcolor{gray!10} \multicolumn{15}{c}{Closed-source MLLMs} \\
\midrule
GPT-4o & 17.2 & 26.4 & 23.2 & 10.8 & 10.6 & 20.3 & 16.1 & 0.7 & 26.9 & 8.5 & 19.2 & 6.4 & 35.1 & 20.3\\
Gemini-2.5-Pro & \underline{30.3} & 44.2 & 22.3 & 25.4 & 34.2 & 24.6 & \underline{31.3} & \underline{12.3} & \underline{42.5} & \underline{20.3} & 28.1 & \underline{10.7} & \underline{58.4} & 30.7\\
Claude-3.5 & 15.4 & 21.3 & 22.5 & 14.4 & 16.2 & 16.3 & 12.2 & 2.1 & 30.5 & 2.2 & 15.3 & 5.1 & 31.4 & 15.4\\
\midrule

\rowcolor{gray!10} \multicolumn{15}{c}{Open-source MLLMs} \\
\midrule
Qwen2.5-VL-72B & 20.2 & 37.1 & 15.4 & 33.2 & 24.3 & \textbf{35.1} & 23.5 & 3.3 & 32.3 & 13.1 & 14.2 & 0.2 & 26.3 & 12.3 \\
Qwen2.5-VL-32B & 19.3 & 26.4 & 19.3 & 25.3 & 10.4 & 18.2 & 23.5 & 2.1 & 14.2 & 15.3 & 13.4 & 5.2 & 48.2 & 13.1\\
Qwen2.5-VL-7B & 16.2 & 21.1 & 11.4 & 48.3 & 14.3 & 15.4 & 24.2 & 0.1 & 22.1 & 11.2 & 24.3 & 0.1 & 21.2 & 9.2\\
LLaVA-Next-72B & 11.3 & 20.3 & 16.1 & 3.2 & 8.2 & 11.2 & 15.3 & 0.1 & 10.1 & 11.2 & 16.3 & 0.1 & 23.3 & 12.1\\
LLaVA-1.6-M-7B & 11.2 & 27.1 & 8.3 & 3.1 & 16.1 & 4.1 & 17.2 & 0.0 & 14.2 & 6.2 & 18.1 & 0.1 & 23.4 & 12.1\\
\midrule
\rowcolor{gray!10} \multicolumn{15}{c}{General Training Models} \\
\midrule
DeepEyes-7B & 17.3 & 22.0 & 6.7 & 41.7 & 19.9 & 16.7 & 19.8 & 1.2 & 16.0 & 3.8 & 19.9 & 3.9 & 50.8 & 12.0  \\
DeepSketcher-7B  & 17.8 & 22.0 & 7.0 & 42.0 & 20.5 & 17.0 & 24.0 & 1.1 & 15.9 & 4.0 & 20.5 & 4.1 & 51.0 & 12.3 \\

\rowcolor{gray!10} \multicolumn{15}{c}{Multimodal Reasoning Frameworks} \\
\midrule
GPT-4o + Sketchpad & 29.8 & \underline{51.2} & \textbf{27.5} & \underline{61.0} & \textbf{55.6} & 16.3 & 24.9 & 9.2 & 33.1 & 18.4 & \textbf{35.1} & 8.1 & 56.8 & \textbf{45.3}\\
GPT-4o + PyVision & 29.2 &\textbf{ 52.1 }& \underline{25.9} & 60.2 & \underline{53.1} & 15.1 & 24.3 & 9.5 & 31.0 & 17.0 & \underline{34.8} & 7.9 & 54.2 & \underline{44.0}  \\
\rowcolor{lavenderDeep}GPT-4o + Octopus & \textbf{33.4} & 47.7 & 25.2 & \textbf{68.5} & 48.8 & \underline{27.6} & \textbf{34.1} & \textbf{15.6} & \textbf{45.2} & \textbf{23.7} & 31.2 & \textbf{13.3} & \textbf{61.6} & 33.8 \\
\midrule
\end{tabular}
}
\caption{
Results on \textbf{Octopus-TIR}, which evaluates multimodal models on 13 visual reasoning and perception tasks (Color, Pro, OCR, SR, Maze, Math, WS, LL-VQA, IR, SD, JG, VS, RG). 
We report accuracy (\%) for closed-source MLLMs, open-source MLLMs, multimodal reasoning frameworks, and generally trained models, with \textbf{All} denoting the average over all tasks.
}\label{tab:TIR_performance}
\end{table*}

\subsection{Main Results}
\begin{figure}
    \centering
    \includegraphics[width=1\linewidth]{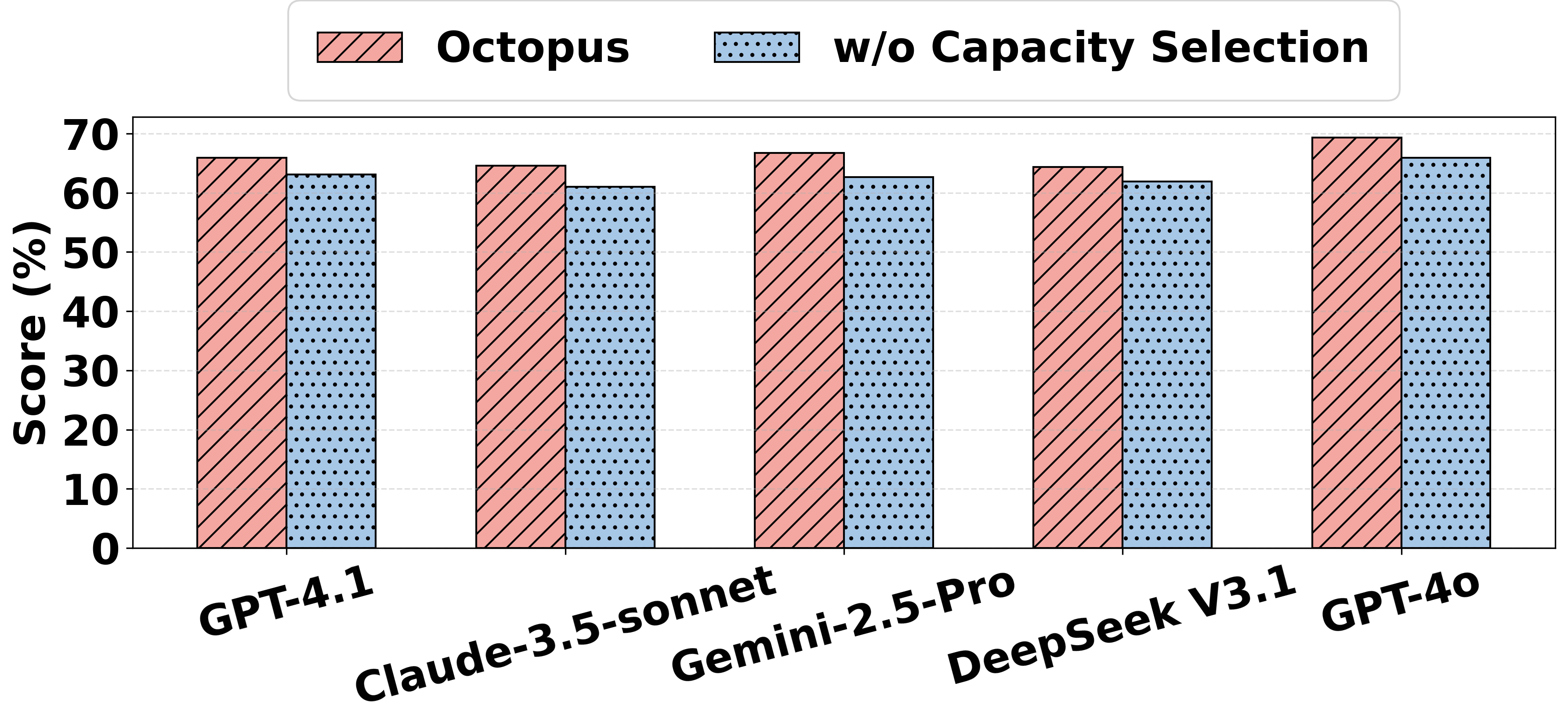}
    \caption{Ablation study on the capacity selection mechanism across five different language models.}
    \label{fig:sense_and_wo_cap}
\end{figure}

We systematically evaluate the proposed Octopus paradigm on Octopus-Bench and its constituent sub-benchmarks. As shown in \Cref{tab:BLINK_performance_comparison}, on the Octopus-BLINK benchmark, our method achieves the best overall performance to date: across 14 fine-grained visual perception and reasoning subtasks, GPT-4o+Octopus obtains an average accuracy of 71.8\%, significantly outperforming existing multimodal frameworks and general-purpose MLLMs. Compared with the strongest baseline (GPT-4o+MMFactory, 68.86\%), our approach achieves a +2.94\% improvement. On Octopus-TIR(\Cref{tab:TIR_performance}), our method again delivers the best overall result: GPT-4o+Octopus reaches 33.4\% on the All metric, surpassing all closed-source/open-source models and multimodal reasoning frameworks. Relative to GPT-4o+Sketchpad (29.8\%) and GPT-4o+PyVision (29.2\%), our approach improves performance by +3.6 and +4.2 \%, respectively. On Octopus-Math benchmark (\Cref{tab:math_performance_comparison}), Octopus also achieves strong performance across all six datasets, including IsoBench (79.2), Geometry3K (48.2), and MathVista (75.3). Furthermore, \Cref{fig:capability_comparison} presents a comparison across the six capability dimensions on Octopus-Bench, showing that Octopus exhibits a more balanced and robust performance than other models. Overall, our capability-first design and structured tool planning consistently enhance diverse multimodal tasks, demonstrating the effectiveness and generality of our unified capability orchestration.

\subsection{Additional Study}
\begin{table}[t]  
\centering
\resizebox{\columnwidth}{!}{
\begin{tabular}{l|cccccc}
\toprule
\textbf{Model} & \textbf{Isobench} & \textbf{Geometry3K} & \textbf{Mathverse} & \textbf{Wemath} & \textbf{Mathvista} & \textbf{Mathvision} \\
\midrule
GPT-4o & 77.5 & 20.1 & 42.1 & 39.2 & 49.1 & 55.5 \\
Gemini-2.5-Pro & \underline{78.2} & 26.3 & 44.3 & \underline{41.6} & 52.2 & \underline{62.3} \\
Claude-3.5 & 70.3 & 37.2 & 41.5 & 40.9 & 42.1 & 53.7 \\
Qwen2.5-VL-72B & 69.3 & 19.2 & 39.2 & 41.5 & 33.8 & 35.6 \\
Qwen2.5-VL-32B & 69.7 & 14.3 & 31.4 & 36.6 & 29.3 & 32.5 \\
Qwen2.5-VL-7B & 50.3 & 8.4 & 29.9 & 34.6 & 23.1 & 27.2 \\
LLaVA-Next-72B & 46.2 & 5.6 & 21.4 & 27.9 & 19.6 & 25.1 \\
LLaVA-1.6-M-7B & 51.3 & 16.5 & 38.8 & 37.9 & 34.2 & 31.4 \\
VILASR-7B ~\cite{wu2025reinforcingspatialreasoningvisionlanguage}& - & - & 29.4 & 23.7 & 57.6 & 25.0 \\
DeepEyes-7B & 25.1 & \underline{46.3} & 42.2 & 38.9 & \underline{70.1} & 26.6 \\
Bagel-Zebra-CoT-7B ~\cite{li2025zebracotdatasetinterleavedvision}& - & - & \underline{48.8} & 28.0 & 64.7 & 28.2 \\
Mirage-7B~\cite{MMI_FrozenLake} & - & - & 27.3 & 16.7 & 63.7 & 28.6 \\
DeepSketcher-7B & 42.1 & 42.7 & 43.2 & 37.1 & 69.1 & 32.3 \\
\rowcolor{lavenderDeep} GPT-4o+Octopus & \textbf{79.2} & \textbf{48.2} & \textbf{49.2} & \textbf{43.1} & \textbf{75.3} & \textbf{65.4} \\
\bottomrule
\end{tabular}
}
\caption{
Results on \textbf{Octopus-Math}.
We report accuracy (\%) on IsoBench~\cite{fu2024isobench}, Geometry3K~\cite{lu2021intergps}, MathVerse~\cite{zhang2024mathverse}, WeMath~\cite{qiao2024wemath}, MathVista~\cite{lu2023mathvista}, and Math-Vision~\cite{wang2024mathvision}.
GPT-4o + Octopus denotes our capability-orchestrated framework built on GPT-4o.
}
\label{tab:math_performance_comparison}
\end{table}



\begin{figure*}[t!]
    \centering
    \includegraphics[width=0.85\textwidth]{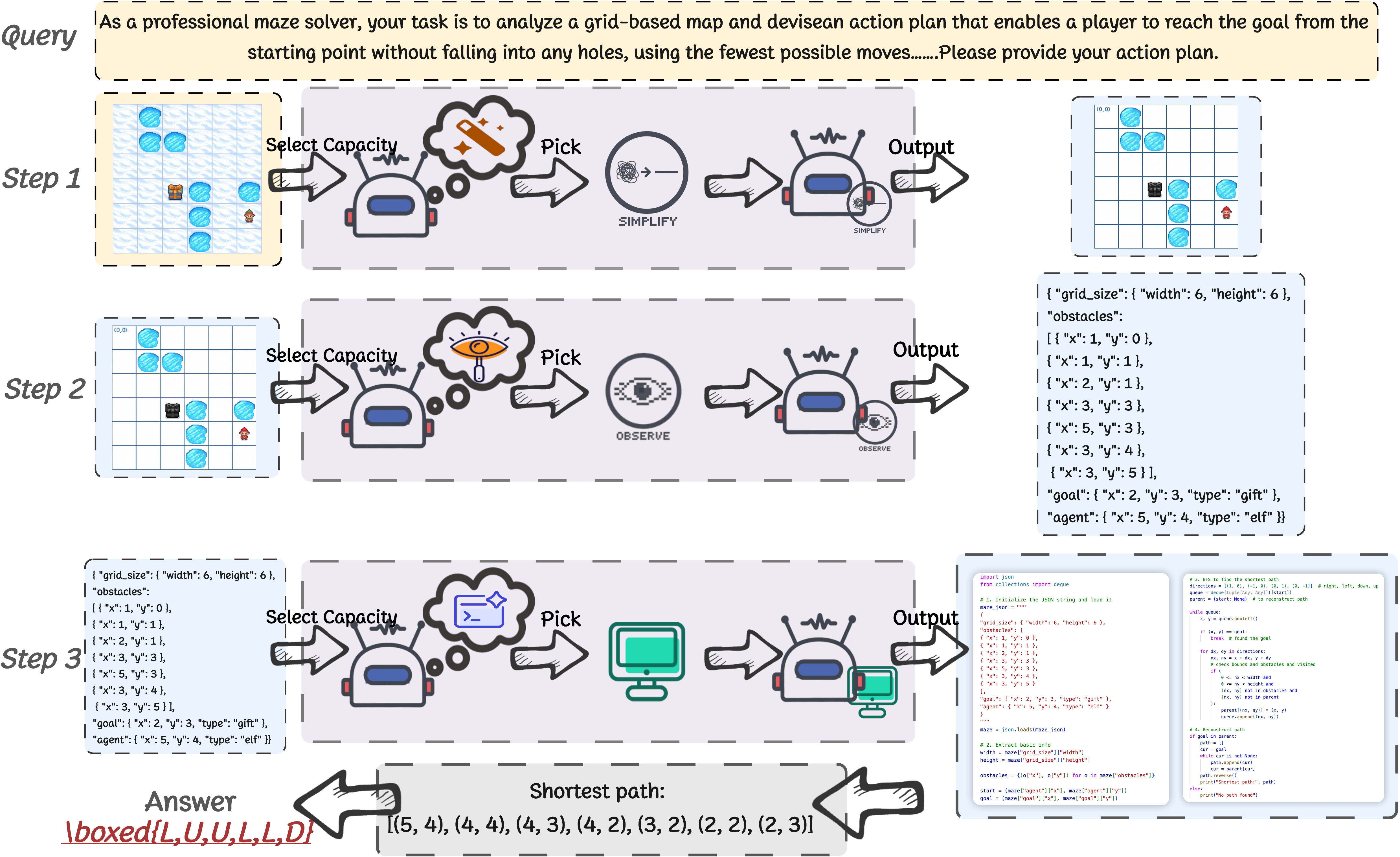}
    \caption{Case study of \textbf{Octopus}. In a visual maze-solving task, Octopus follows a capability-first workflow: Step~1 simplifies the raw map, Step~2 extracts grid-level semantics, and Step~3 computes the optimal path through logical reasoning. The final solution emerges from integrating multimodal evidence across these steps.}
    \label{fig:navi-case}
\end{figure*}

\subsubsection{Necessity of the Six Capabilities}
Octopus highlights the importance of comprehensive capability coverage in multimodal reasoning. As shown in Figure~\ref{fig:ablation_cap}, on both Octopus-BLINK and Octopus-TIR, removing any single capability leads to a noticeable performance drop of approximately 5–10\%. An interesting observation is that removing the logic capability causes the most severe degradation. This suggests that when tasks involve complex reasoning patterns that cannot be resolved through simple combinations of existing visual tools, the logic capability enables Octopus to perform rigorous analysis in the computation and code space, thereby playing a critical role in achieving high-quality reasoning.

Moreover, as shown in Figure~\ref{fig:sense_and_wo_cap}, when we disable capability selection and allow the agent to choose tools directly from the full tool set, performance drops to a certain extent. This indicates that the capability-first strategy effectively mimics how humans decompose tasks in the reasoning space, leading to more stable and structurally coherent decision-making.
\subsubsection{Sensitive Study}

We additionally evaluate how different MLLMs perform when used as the backbone of our unified model. As shown in Figure~\ref{fig:sense_and_wo_cap}, Octopus exhibits consistently stable performance across various reasoning backbone model choices, demonstrating the robustness of our method with respect to the underlying MLLM. 
\subsection{Case Study}
Figure~\ref{fig:navi-case} illustrates a representative example from Octopus-Bench, where the Octopus is asked to solve a grid-based maze and produce the shortest safe action sequence. The agent progressively selects different capabilities to decompose the task. The process begins with the Generate capability, which is used to transform the raw maze image into a clean and structured grid representation. Then, using fine-grained perception, the agent extracts structured information such as obstacles, the agent’s location, and the goal position. Finally, the agent invokes logical programming reasoning to compute the shortest collision-free path through code execution, yielding a coordinate-level trajectory.

Based on this computed path, the agent generates the final action plan, producing the optimal movement sequence (e.g. L,U,U,L,L,L,D). This case demonstrates how Octopus integrates different capabilities to solve multi-step navigation tasks in a mutlimodal reasoning space. 
\section{Conclusion}

In this work, we propose Octopus, a new agentic reasoning paradigm for multimodal reasoning tasks. We systematically define six fundamental capabilities essential for multimodal reasoning and further construct a comprehensive evaluation benchmark—Octopus-Bench—to assess the capability coverage of existing MLLMs and agent frameworks. Our framework enables models to autonomously select the most appropriate capability at each reasoning step, forming human-like reasoning trajectories through dynamic composition and adaptive decision-making. Experimental results on Octopus-Bench demonstrate that Octopus significantly outperforms state-of-the-art MLLMs and agent framwork. We call for future research in multimodal reasoning to move beyond tool-driven approaches toward capability-level coordination, paving the way for next-generation intelligent systems that are more generalizable, interpretable, and autonomous.
{
    \small
    \bibliographystyle{ieeenat_fullname}
    \bibliography{main}

@String(CVPR= {IEEE Conf. Comput. Vis. Pattern Recog.})

@String(AAAI = {AAAI})

@String(CVPR  = {CVPR})

@misc{li2025searcho1agenticsearchenhancedlarge,
      title={Search-o1: Agentic Search-Enhanced Large Reasoning Models}, 
      author={Xiaoxi Li and Guanting Dong and Jiajie Jin and Yuyao Zhang and Yujia Zhou and Yutao Zhu and Peitian Zhang and Zhicheng Dou},
      year={2025},
      eprint={2501.05366},
      archivePrefix={arXiv},
      primaryClass={cs.AI},
      url={https://arxiv.org/abs/2501.05366}, 
}

@misc{yang2023dawnlmmspreliminaryexplorations,
      title={The Dawn of LMMs: Preliminary Explorations with GPT-4V(ision)}, 
      author={Zhengyuan Yang and Linjie Li and Kevin Lin and Jianfeng Wang and Chung-Ching Lin and Zicheng Liu and Lijuan Wang},
      year={2023},
      eprint={2309.17421},
      archivePrefix={arXiv},
      primaryClass={cs.CV},
      url={https://arxiv.org/abs/2309.17421}, 
}

@article{comanici2025gemini,
  title={Gemini 2.5: Pushing the frontier with advanced reasoning, multimodality, long context, and next generation agentic capabilities},
  author={Comanici, Gheorghe and Bieber, Eric and Schaekermann, Mike and Pasupat, Ice and Sachdeva, Noveen and Dhillon, Inderjit and Blistein, Marcel and Ram, Ori and Zhang, Dan and Rosen, Evan and others},
  journal={arXiv preprint arXiv:2507.06261},
  year={2025}
}

@article{bai2025qwen2,
  title={Qwen2. 5-vl technical report},
  author={Bai, Shuai and Chen, Keqin and Liu, Xuejing and Wang, Jialin and Ge, Wenbin and Song, Sibo and Dang, Kai and Wang, Peng and Wang, Shijie and Tang, Jun and others},
  journal={arXiv preprint arXiv:2502.13923},
  year={2025}
}

@inproceedings{chen2024internvl,
  title={Internvl: Scaling up vision foundation models and aligning for generic visual-linguistic tasks},
  author={Chen, Zhe and Wu, Jiannan and Wang, Wenhai and Su, Weijie and Chen, Guo and Xing, Sen and Zhong, Muyan and Zhang, Qinglong and Zhu, Xizhou and Lu, Lewei and others},
  booktitle={Proceedings of the IEEE/CVF conference on computer vision and pattern recognition},
  pages={24185--24198},
  year={2024}
}

@article{gao2024mini,
  title={Mini-internvl: a flexible-transfer pocket multi-modal model with 5\% parameters and 90\% performance},
  author={Gao, Zhangwei and Chen, Zhe and Cui, Erfei and Ren, Yiming and Wang, Weiyun and Zhu, Jinguo and Tian, Hao and Ye, Shenglong and He, Junjun and Zhu, Xizhou and others},
  journal={Visual Intelligence},
  volume={2},
  number={1},
  pages={32},
  year={2024},
  publisher={Springer}
}

@article{wei2022chain,
  title={Chain-of-thought prompting elicits reasoning in large language models},
  author={Wei, Jason and Wang, Xuezhi and Schuurmans, Dale and Bosma, Maarten and Xia, Fei and Chi, Ed and Le, Quoc V and Zhou, Denny and others},
  journal={Advances in neural information processing systems},
  volume={35},
  pages={24824--24837},
  year={2022}
}

@article{zhang2023multimodal,
  title={Multimodal chain-of-thought reasoning in language models},
  author={Zhang, Zhuosheng and Zhang, Aston and Li, Mu and Zhao, Hai and Karypis, George and Smola, Alex},
  journal={arXiv preprint arXiv:2302.00923},
  year={2023}
}

@inproceedings{yao2022react,
  title={React: Synergizing reasoning and acting in language models},
  author={Yao, Shunyu and Zhao, Jeffrey and Yu, Dian and Du, Nan and Shafran, Izhak and Narasimhan, Karthik R and Cao, Yuan},
  booktitle={The eleventh international conference on learning representations},
  year={2022}
}

@article{hu2024visual,
  title={Visual sketchpad: Sketching as a visual chain of thought for multimodal language models},
  author={Hu, Yushi and Shi, Weijia and Fu, Xingyu and Roth, Dan and Ostendorf, Mari and Zettlemoyer, Luke and Smith, Noah A and Krishna, Ranjay},
  journal={Advances in Neural Information Processing Systems},
  volume={37},
  pages={139348--139379},
  year={2024}
}

@inproceedings{wu2024dettoolchain,
  title={Dettoolchain: A new prompting paradigm to unleash detection ability of mllm},
  author={Wu, Yixuan and Wang, Yizhou and Tang, Shixiang and Wu, Wenhao and He, Tong and Ouyang, Wanli and Torr, Philip and Wu, Jian},
  booktitle={European Conference on Computer Vision},
  pages={164--182},
  year={2024},
  organization={Springer}
}

@article{zhou2024image,
  title={Image-of-thought prompting for visual reasoning refinement in multimodal large language models},
  author={Zhou, Qiji and Zhou, Ruochen and Hu, Zike and Lu, Panzhong and Gao, Siyang and Zhang, Yue},
  journal={arXiv preprint arXiv:2405.13872},
  year={2024}
}

@inproceedings{suris2023vipergpt,
  title={Vipergpt: Visual inference via python execution for reasoning},
  author={Sur{\'\i}s, D{\'\i}dac and Menon, Sachit and Vondrick, Carl},
  booktitle={Proceedings of the IEEE/CVF international conference on computer vision},
  pages={11888--11898},
  year={2023}
}

@InProceedings{Rodriguez_2025_CVPR,
    author    = {Rodriguez, Juan A. and Puri, Abhay and Agarwal, Shubham and Laradji, Issam H. and Rodriguez, Pau and Rajeswar, Sai and Vazquez, David and Pal, Christopher and Pedersoli, Marco},
    title     = {StarVector: Generating Scalable Vector Graphics Code from Images and Text},
    booktitle = {Proceedings of the IEEE/CVF Conference on Computer Vision and Pattern Recognition (CVPR)},
    month     = {June},
    year      = {2025},
    pages     = {16175-16186}
}

@inproceedings{awal2025webmmu,
  title={Webmmu: A benchmark for multimodal multilingual website understanding and code generation},
  author={Awal, Rabiul and Massoud, Mahsa and Feizi, Aarash and Li, Zichao and Wang, Suyuchen and Pal, Christopher and Agrawal, Aishwarya and Vazquez, David and Reddy, Siva and Rodriguez, Juan A and others},
  booktitle={Proceedings of the 2025 Conference on Empirical Methods in Natural Language Processing},
  pages={25129--25156},
  year={2025}
}

@article{chern2025thinking,
  title={Thinking with Generated Images},
  author={Chern, Ethan and Hu, Zhulin and Chern, Steffi and Kou, Siqi and Su, Jiadi and Ma, Yan and Deng, Zhijie and Liu, Pengfei},
  journal={arXiv preprint arXiv:2505.22525},
  year={2025}
}

@article{xu2025visual,
  title={Visual Planning: Let's Think Only with Images},
  author={Xu, Yi and Li, Chengzu and Zhou, Han and Wan, Xingchen and Zhang, Caiqi and Korhonen, Anna and Vuli{\'c}, Ivan},
  journal={arXiv preprint arXiv:2505.11409},
  year={2025}
}

@article{wu2024mind,
  title={Mind's eye of LLMs: visualization-of-thought elicits spatial reasoning in large language models},
  author={Wu, Wenshan and Mao, Shaoguang and Zhang, Yadong and Xia, Yan and Dong, Li and Cui, Lei and Wei, Furu},
  journal={Advances in Neural Information Processing Systems},
  volume={37},
  pages={90277--90317},
  year={2024}
}

@misc{li2024llavaonevisioneasyvisualtask,
      title={LLaVA-OneVision: Easy Visual Task Transfer}, 
      author={Bo Li and Yuanhan Zhang and Dong Guo and Renrui Zhang and Feng Li and Hao Zhang and Kaichen Zhang and Peiyuan Zhang and Yanwei Li and Ziwei Liu and Chunyuan Li},
      year={2024},
      eprint={2408.03326},
      archivePrefix={arXiv},
      primaryClass={cs.CV},
      url={https://arxiv.org/abs/2408.03326}, 
}

@inproceedings{kojima2022large,
  author    = {Kojima, Takeshi and Gu, Shixiang Shane and Reid, Machel and Matsuo, Yutaka and Iwasawa, Yusuke},
  title     = {Large Language Models are Zero-Shot Reasoners},
  booktitle = {Advances in Neural Information Processing Systems},
  volume    = {35},
  pages     = {22199--22213},
  year      = {2022},
  url       = {https://arxiv.org/abs/2205.11916}
}

@article{openai2024o1,
  author    = {OpenAI},
  title     = {OpenAI o1 System Card},
  journal   = {arXiv preprint arXiv:2412.16720},
  year      = {2024},
  url       = {https://arxiv.org/abs/2412.16720}
}

@article{liu2023visual,
  author    = {Liu, Haotian and Li, Chunyuan and Wu, Qingyang and Lee, Yong Jae},
  title     = {Visual Instruction Tuning},
  journal   = {Advances in Neural Information Processing Systems},
  volume    = {36},
  year      = {2023},
  url       = {https://arxiv.org/abs/2304.08485}
}

@article{zhu2023minigpt,
  author    = {Zhu, Deyao and Chen, Jun and Shen, Xiaoqian and Li, Xiang and Elhoseiny, Mohamed},
  title     = {MiniGPT-4: Enhancing Vision-Language Understanding with Advanced Large Language Models},
  journal   = {arXiv preprint arXiv:2304.10592},
  year      = {2023},
  url       = {https://arxiv.org/abs/2304.10592}
}

@article{chen2023shikra,
  author    = {Chen, Hao and Zhang, Anjiao and Yao, Zhiyu and Yang, Shuai and Zhao, Jiawei and Yu, Jie and Zhang, Kekai and Li, Q. and Sun, Zhao and Jia, Jia},
  title     = {Shikra: Unleashing Multimodal LLM's Referential Dialogue Magic},
  journal   = {arXiv preprint arXiv:2306.15195},
  year      = {2023},
  url       = {https://arxiv.org/abs/2306.15195}
}

@article{kunda2013computational,
  title={A computational model for solving problems from the Raven's Progressive Matrices intelligence test using iconic visual representations},
  author={Kunda, Maithilee and McGreggor, Keith and Goel, Ashok K},
  journal={Cognitive Systems Research},
  volume={22},
  pages={47--66},
  year={2013},
  publisher={Elsevier}
}

@article{vaishnav2025cognitive,
  title={Cognitive Paradigms for Evaluating VLMs on Visual Reasoning Task},
  author={Vaishnav, Mihir and Tammet, Tanel},
  journal={arXiv preprint arXiv:2501.13620},
  year={2025}
}

@misc{fan2024mmfactoryuniversalsolutionsearch,
      title={MMFactory: A Universal Solution Search Engine for Vision-Language Tasks}, 
      author={Wan-Cyuan Fan and Tanzila Rahman and Leonid Sigal},
      year={2024},
      eprint={2412.18072},
      archivePrefix={arXiv},
      primaryClass={cs.CV},
      url={https://arxiv.org/abs/2412.18072}, 
}

@article{achiam2023gpt,
  title={Gpt-4 technical report},
  author={Achiam, Josh and Adler, Steven and Agarwal, Sandhini and Ahmad, Lama and Akkaya, Ilge and Aleman, Florencia Leoni and Almeida, Diogo and Altenschmidt, Janko and Altman, Sam and Anadkat, Shyamal and others},
  journal={arXiv preprint arXiv:2303.08774},
  year={2023}
}

@online{anthropic2024claude35sonnet,
  author       = {Anthropic PBC},
  title        = {Claude 3.5 Sonnet},
  year         = {2024},
  month        = jun,
  day          = {21},
  url          = {https://www.anthropic.com/news/claude-3-5-sonnet},
  note         = {Accessed: 2025-11-13}
}

@misc{zheng2025deepeyesincentivizingthinkingimages,
      title={DeepEyes: Incentivizing "Thinking with Images" via Reinforcement Learning}, 
      author={Ziwei Zheng and Michael Yang and Jack Hong and Chenxiao Zhao and Guohai Xu and Le Yang and Chao Shen and Xing Yu},
      year={2025},
      eprint={2505.14362},
      archivePrefix={arXiv},
      primaryClass={cs.CV},
      url={https://arxiv.org/abs/2505.14362}, 
}

@misc{zhang2025deepsketcherinternalizingvisualmanipulation,
      title={DeepSketcher: Internalizing Visual Manipulation for Multimodal Reasoning}, 
      author={Chi Zhang and Haibo Qiu and Qiming Zhang and Zhixiong Zeng and Lin Ma and Jing Zhang},
      year={2025},
      eprint={2509.25866},
      archivePrefix={arXiv},
      primaryClass={cs.CV},
      url={https://arxiv.org/abs/2509.25866}, 
}

@inproceedings{blink,
  title={Blink: Multimodal large language models can see but not perceive},
  author={Fu, Xingyu and Hu, Yushi and Li, Bangzheng and Feng, Yu and Wang, Haoyu and Lin, Xudong and Roth, Dan and Smith, Noah A and Ma, Wei-Chiu and Krishna, Ranjay},
  booktitle={European Conference on Computer Vision},
  pages={148--166},
  year={2024},
  organization={Springer}
}

@misc{tirbench,
      title={TIR-Bench: A Comprehensive Benchmark for Agentic Thinking-with-Images Reasoning}, 
      author={Ming Li and Jike Zhong and Shitian Zhao and Haoquan Zhang and Shaoheng Lin and Yuxiang Lai and Chen Wei and Konstantinos Psounis and Kaipeng Zhang},
      year={2025},
      eprint={2511.01833},
      archivePrefix={arXiv},
      primaryClass={cs.CV},
      url={https://arxiv.org/abs/2511.01833}, 
}

@inproceedings{comt,
  title={Comt: A novel benchmark for chain of multi-modal thought on large vision-language models},
  author={Cheng, Zihui and Chen, Qiguang and Zhang, Jin and Fei, Hao and Feng, Xiaocheng and Che, Wanxiang and Li, Min and Qin, Libo},
  booktitle={Proceedings of the AAAI Conference on Artificial Intelligence},
  volume={39},
  number={22},
  pages={23678--23686},
  year={2025}
}

@inproceedings{vstar,
  title={V?: Guided visual search as a core mechanism in multimodal llms},
  author={Wu, Penghao and Xie, Saining},
  booktitle={Proceedings of the IEEE/CVF Conference on Computer Vision and Pattern Recognition},
  pages={13084--13094},
  year={2024}
}

@inproceedings{mmvp,
  title={Eyes wide shut? exploring the visual shortcomings of multimodal llms},
  author={Tong, Shengbang and Liu, Zhuang and Zhai, Yuexiang and Ma, Yi and LeCun, Yann and Xie, Saining},
  booktitle={Proceedings of the IEEE/CVF Conference on Computer Vision and Pattern Recognition},
  pages={9568--9578},
  year={2024}
}

@article{MMI_FrozenLake,
  title={Machine Mental Imagery: Empower Multimodal Reasoning with Latent Visual Tokens},
  author={Yang, Zeyuan and Yu, Xueyang and Chen, Delin and Shen, Maohao and Gan, Chuang},
  journal={arXiv preprint arXiv:2506.17218},
  year={2025}
}

@misc{fu2024isobench,
  author = {Deqing Fu and Ghazal Khalighinejad and Ollie Liu and Bhuwan Dhingra and Dani Yogatama and Robin Jia and Willie Neiswanger},
  title = {IsoBench: Benchmarking Multimodal Foundation Models on Isomorphic Representations},
  archivePrefix = {arXiv},
  eprint = {2404.01266},
  primaryClass = {cs.AI},
  year = {2024}
}

@misc{lu2021intergps,
  author = {Pan Lu and Ran Gong and Shibiao Jiang and Liang Qiu and Siyuan Huang and Xiaodan Liang and Song-Chun Zhu},
  title = {Inter-GPS: Interpretable Geometry Problem Solving with Formal Language and Symbolic Reasoning},
  archivePrefix = {arXiv},
  eprint = {2105.04165},
  primaryClass = {cs.CL},
  year = {2021}
}

@misc{zhang2024mathverse,
  author = {Renrui Zhang and Dongzhi Jiang and Yichi Zhang and Haokun Lin and Z. J. Guo and Pengshuo Qiu and Aojun Zhou and Pan Lu and Kai-Wei Chang and Peng Gao and Hongsheng Li},
  title = {MathVerse: Does Your Multi-modal LLM Truly See the Diagrams in Visual Math Problems?},
  archivePrefix = {arXiv},
  eprint = {2403.14624},
  primaryClass = {cs.CV},
  year = {2024}
}

@misc{qiao2024wemath,
  author = {Runqi Qiao and Qiuna Tan and Guanting Dong and Minhui Wu and Chong Sun and Xiaoshuai Song and Zhuoma GongQue and Shanglin Lei and Zhe Wei and Miaoxuan Zhang and Runfeng Qiao and Yifan Zhang and Xiao Zong and Yida Xu and Muxi Diao and Zhimin Bao and Chen Li and Honggang Zhang},
  title = {We-Math: Does Your Large Multimodal Model Achieve Human-like Mathematical Reasoning?},
  archivePrefix = {arXiv},
  eprint = {2407.01284},
  primaryClass = {cs.AI},
  year = {2024}
}

@misc{lu2023mathvista,
  author = {Pan Lu and Hritik Bansal and Tony Xia and Jiacheng Liu and Chunyuan Li and Hannaneh Hajishirzi and Hao Cheng and Kai-Wei Chang and Michel Galley and Jianfeng Gao},
  title = {MathVista: Evaluating Mathematical Reasoning of Foundation Models in Visual Contexts},
  archivePrefix = {arXiv},
  eprint = {2310.02255},
  primaryClass = {cs.CV},
  year = {2023}
}

@misc{wang2024mathvision,
  author = {Ke Wang and Junting Pan and Weikang Shi and Zimu Lu and Mingjie Zhan and Hongsheng Li},
  title = {Measuring Multimodal Mathematical Reasoning with MATH-Vision Dataset},
  archivePrefix = {arXiv},
  eprint = {2402.14804},
  primaryClass = {cs.CV},
  year = {2024}
}

@misc{openai2024gpt4o,
  author = {OpenAI and Hurst, Aaron and others},
  title = {GPT-4o System Card},
  eprint = {2410.21276},
  archivePrefix = {arXiv},
  primaryClass = {cs.CL},
  url = {https://arxiv.org/abs/2410.21276},
  doi = {10.48550/arXiv.2410.21276},
  month = {10},
  year = {2024}
}

@article{VTS-V,
  title={Multi-Step Visual Reasoning with Visual Tokens Scaling and Verification},
  author={Bai, Tianyi and Hu, Zengjie and Sun, Fupeng and Qiu, Jiantao and Jiang, Yizhen and He, Guangxin and Zeng, Bohan and He, Conghui and Yuan, Binhang and Zhang, Wentao},
  journal={arXiv preprint arXiv:2506.07235},
  year={2025}
}

@article{team2025gemini, author = {gemini team}, title = {Gemini 2.5: Pushing the Frontier with Advanced Reasoning, Multimodality, Long Context, and Next Generation Agentic Capabilities}, journal = {arXiv}, year = {2025}, month = {July}, url = {https://arxiv.org/abs/2507.06261}, doi = {10.48550/arXiv.2507.06261} }

@misc{anthropic2025claudesonnet, author = {Anthropic}, title = {Claude Sonnet 4.5 System Card}, year = {2025}, url = {https://www.anthropic.com/claude-sonnet-4-5-system-card}, note = {System card} }

@misc{liu2023llavapluslearningusetools,
      title={LLaVA-Plus: Learning to Use Tools for Creating Multimodal Agents}, 
      author={Shilong Liu and Hao Cheng and Haotian Liu and Hao Zhang and Feng Li and Tianhe Ren and Xueyan Zou and Jianwei Yang and Hang Su and Jun Zhu and Lei Zhang and Jianfeng Gao and Chunyuan Li},
      year={2023},
      eprint={2311.05437},
      archivePrefix={arXiv},
      primaryClass={cs.CV},
      url={https://arxiv.org/abs/2311.05437}, 
}

@misc{zhao2025pyvisionagenticvisiondynamic,
      title={PyVision: Agentic Vision with Dynamic Tooling}, 
      author={Shitian Zhao and Haoquan Zhang and Shaoheng Lin and Ming Li and Qilong Wu and Kaipeng Zhang and Chen Wei},
      year={2025},
      eprint={2507.07998},
      archivePrefix={arXiv},
      primaryClass={cs.CL},
      url={https://arxiv.org/abs/2507.07998}, 
}

@misc{li2024llavanext-strong,
    title={LLaVA-NeXT: Stronger LLMs Supercharge Multimodal Capabilities in the Wild},
    url={https://llava-vl.github.io/blog/2024-05-10-llava-next-stronger-llms/},
    author={Li, Bo and Zhang, Kaichen and Zhang, Hao and Guo, Dong and Zhang, Renrui and Li, Feng and Zhang, Yuanhan and Liu, Ziwei and Li, Chunyuan},
    month={May},
    year={2024}
}

@article{wang2024qwen2,
  title={Qwen2-vl: Enhancing vision-language model's perception of the world at any resolution},
  author={Wang, Peng and Bai, Shuai and Tan, Sinan and Wang, Shijie and Fan, Zhihao and Bai, Jinze and Chen, Keqin and Liu, Xuejing and Wang, Jialin and Ge, Wenbin and others},
  journal={arXiv preprint arXiv:2409.12191},
  year={2024}
}

@misc{li2025zebracotdatasetinterleavedvision,
      title={Zebra-CoT: A Dataset for Interleaved Vision Language Reasoning}, 
      author={Ang Li and Charles Wang and Deqing Fu and Kaiyu Yue and Zikui Cai and Wang Bill Zhu and Ollie Liu and Peng Guo and Willie Neiswanger and Furong Huang and Tom Goldstein and Micah Goldblum},
      year={2025},
      eprint={2507.16746},
      archivePrefix={arXiv},
      primaryClass={cs.CV},
      url={https://arxiv.org/abs/2507.16746}, 
}

@misc{yang2023setofmarkpromptingunleashesextraordinary,
      title={Set-of-Mark Prompting Unleashes Extraordinary Visual Grounding in GPT-4V}, 
      author={Jianwei Yang and Hao Zhang and Feng Li and Xueyan Zou and Chunyuan Li and Jianfeng Gao},
      year={2023},
      eprint={2310.11441},
      archivePrefix={arXiv},
      primaryClass={cs.CV},
      url={https://arxiv.org/abs/2310.11441}, 
}

@misc{wu2025reinforcingspatialreasoningvisionlanguage,
      title={Reinforcing Spatial Reasoning in Vision-Language Models with Interwoven Thinking and Visual Drawing}, 
      author={Junfei Wu and Jian Guan and Kaituo Feng and Qiang Liu and Shu Wu and Liang Wang and Wei Wu and Tieniu Tan},
      year={2025},
      eprint={2506.09965},
      archivePrefix={arXiv},
      primaryClass={cs.CV},
      url={https://arxiv.org/abs/2506.09965}, 
}

@misc{seagent,
      title={SE-Agent: Self-Evolution Trajectory Optimization in Multi-Step Reasoning with LLM-Based Agents}, 
      author={Jiaye Lin and Yifu Guo and Yuzhen Han and Sen Hu and Ziyi Ni and Licheng Wang and Mingguang Chen and Hongzhang Liu and Ronghao Chen and Yangfan He and Daxin Jiang and Binxing Jiao and Chen Hu and Huacan Wang},
      year={2025},
      eprint={2508.02085},
      archivePrefix={arXiv},
      primaryClass={cs.AI},
      url={https://arxiv.org/abs/2508.02085}, 
}

@misc{decoupleq,
      title={Decoupling Continual Semantic Segmentation}, 
      author={Yifu Guo and Yuquan Lu and Wentao Zhang and Zishan Xu and Dexia Chen and Siyu Zhang and Yizhe Zhang and Ruixuan Wang},
      year={2025},
      eprint={2508.05065},
      archivePrefix={arXiv},
      primaryClass={cs.CV},
      url={https://arxiv.org/abs/2508.05065}, 
}

@inproceedings{du2025mokgr,
  author    = {Du, Enjun and Liu, Siyi and Zhang, Yongqi},
  title     = {Mixture of Length and Pruning Experts for Knowledge Graphs Reasoning},
  booktitle = {Proceedings of the 2025 Conference on Empirical Methods in Natural Language Processing (EMNLP 2025)},
  pages     = {432--453},
  year      = {2025},
  month     = nov
}

@inproceedings{du2025graphmaster,
  author    = {Du, Enjun and Li, Xunkai and Jin, Tian and Zhang, Zhihan and Li, Rong{-}Hua and Wang, Guoren},
  title     = {GraphMaster: Automated Graph Synthesis via {LLM} Agents in Data-Limited Environments},
  booktitle = {Advances in Neural Information Processing Systems 39 (NeurIPS 2025)},
  year      = {2025},
  month     = apr
}

@article{du2025graphoracle,
  title={GraphOracle: A Foundation Model for Knowledge Graph Reasoning},
  author={Du, Enjun and Liu, Siyu and Zhang, Yongqi},
  journal={arXiv preprint arXiv:2505.11125},
  year={2025},
  url={https://arxiv.org/abs/2505.11125}
}

@misc{wang2025repomasterautonomousexplorationunderstanding,
      title={RepoMaster: Autonomous Exploration and Understanding of GitHub Repositories for Complex Task Solving}, 
      author={Huacan Wang and Ziyi Ni and Shuo Zhang and Shuo Lu and Sen Hu and Ziyang He and Chen Hu and Jiaye Lin and Yifu Guo and Ronghao Chen and Xin Li and Daxin Jiang and Yuntao Du and Pin Lyu},
      year={2025},
      eprint={2505.21577},
      archivePrefix={arXiv},
      primaryClass={cs.SE},
      url={https://arxiv.org/abs/2505.21577}, 
}

@misc{luo2025codetestcasesenough,
      title={How Many Code and Test Cases Are Enough? Evaluating Test Cases Generation from a Binary-Matrix Perspective}, 
      author={Xianzhen Luo and Jinyang Huang and Wenzhen Zheng and Qingfu Zhu and Mingzheng Xu and Yiheng Xu and Yuantao Fan and Libo Qin and Wanxiang Che},
      year={2025},
      eprint={2510.08720},
      archivePrefix={arXiv},
      primaryClass={cs.CL},
      url={https://arxiv.org/abs/2510.08720}, 
}
}


\end{document}